\documentclass[10pt,twocolumn,letterpaper]{article}

\usepackage{cvpr}
\usepackage{times}
\usepackage{epsfig}
\usepackage{graphicx}
\usepackage{amsmath}
\usepackage{amssymb}

\usepackage{float}
\usepackage[utf8]{inputenc}
\usepackage{makeidx}
\usepackage{url}
\usepackage{epstopdf}
\usepackage{doc}
\usepackage{enumerate}
\usepackage{bbm}
\usepackage{multicol, blindtext}
\usepackage{ctable}
\usepackage{textcomp}			
\usepackage[singlelinecheck=false]{caption}
\usepackage[]{footmisc}
\usepackage{multirow}
\usepackage{subfigure}
\usepackage{lipsum}
\usepackage{adjustbox}
\usepackage{pgfplots}

\usepackage[font={small}]{caption}
\usepackage{tikz}
\usepackage[pagebackref=true,breaklinks=true,letterpaper=true,colorlinks,bookmarks=false]{hyperref}

\cvprfinalcopy 

\def\cvprPaperID{8038} 

\begin{document}

\title{Unsupervised Monocular Depth Prediction for Indoor Continuous Video Streams}

\author{\parbox{16cm}{\centering
		{Yinglong Feng, \hspace{0.2cm} Shuncheng Wu, \hspace{0.2cm} Okan K\"op\"ukl\"u, \hspace{0.2cm} Xueyang Kang, \hspace{0.2cm} Federico Tombari}\\
		{\normalsize
			\vspace{0.2cm}
			Technical University of Munich, Germany}}
}

\maketitle
\begin{abstract}
This paper studies unsupervised monocular depth prediction problem.  Most of existing unsupervised depth prediction algorithms are developed for outdoor scenarios, while the depth prediction work in the indoor environment is still very scarce to our knowledge. Therefore, this work focuses on narrowing the gap by firstly evaluating existing approaches in the indoor environments and then improving the state-of-the-art design of architecture. Unlike typical outdoor training dataset, such as KITTI with motion constraints, data for indoor environment contains more arbitrary camera movement and short baseline between two consecutive images, which deteriorates the network training for the pose estimation. To address this issue, we propose two methods: 
Firstly, we propose a novel reconstruction loss function to constraint pose estimation, resulting in accuracy improvement of the predicted disparity map; secondly, we use an ensemble learning with a flipping strategy along with a median filter, directly taking operation on the output disparity map. We evaluate our approaches on the TUM RGB-D and self-collected datasets. The results have shown that both approaches outperform the previous \mbox{state-of-the-art} unsupervised learning approaches.
\end{abstract}

\section{Introduction}
At present, development in computer vision is shifting from 2D to 3D. With the advent of the autonomous driving era and with the increasing popularity of head-mounted devices, people are increasingly demanding 3D scene information. 3D perception tasks require depth information in addition to 2D information. With 3D information, further complex tasks can be performed, such as 3D scene understanding or 3D collision avoidance.  The depth information is also vital for the self-driving vehicle to estimate the distance to surrounding objects for safety insurance.

The common depth estimation approach utilizes a stereo camera, with known a baseline configuration. A depth map of the current frame can be computed by using disparity information between left and right images. This approach has been widely used since its basic mechanism is quite similar to the visual perception system of human. However, stereo cameras increase hardware expense and the majority of smartphones are only equipped with a monocular camera to reduce the production cost. From the scientific point of view, monocular video stream is more complicated and challenging in terms of 3D pose estimation and depth prediction.

Generally, supervised depth estimation approaches such as \cite{deeperdepth} requires a large training dataset containing ground truth (GT) depth maps. Although supervised learning can utilize regression methods to initialize a good depth map as an initial point, the GT of depth is difficult to obtain in reality ; hence the supervised learning approach for depth estimation is often impractical. To overcome this issue, the unsupervised \cite{monodepth1} and self-supervised \cite{vid2depth} approaches have been proposed such that depth estimation can be done without acquiring ground truth data. 

In most of the recent works, the unsupervised depth estimation methods are trained and evaluated on outdoor datasets such as KITTI \cite{kitti} and Cityscape \cite{cityscape}. The way of GT acquisition for outdoor scenes is to project lidar information on the two-dimensional plane. However, the scene from these recordings has some limited patterns. For instance, the two sides of the road are populated with trees, buildings, and parking cars, while the upper part of the most images is only filled with open sky. In that case, the depth information in these datasets follows a biased distribution and valid depth information only occurs in the lower part of the image. This also explains why the models trained on the KITTI or CityScapes, do not generalize well on other datasets like TUM RGB-D dataset \cite{tumrgbd}.

In this paper, we initially evaluate the state-of-art unsupervised learning methods in the indoor scenarios, which fail to achieve acceptable performance. Then we propose first a reconstruction loss function to provide additional constraint to the pose estimation, and second ensemble learning with flipping strategy together with median filtering. The proposed approach outperforms previous state-of-the-art results on TUM RGB-D and self-collected X-tion datasets. 


\section{Related Work}
Early works on the monocular depth prediction algorithm are reviewed in this chapter discussing their similarities and differences. There are mainly two types of depth prediction algorithms: Structure from Motion (SFM) and deep learning based depth prediction. 

\subsection{Structure from Motion-based Depth Prediction}

Depth estimation is a traditional 3D vision problem. In \cite{mvg}, the algorithm for estimating the depth from stereo camera, namely semi-global block matching (SGBM) \cite{sgbm}, is described in detail. The multi-view geometry based method is robust and has a good applicability in different scenarios. But the drawback of the traditional approach is that a good initialization is required and the generated depth information is not dense, which means a subsequent reprocessing is required for dense reconstruction. Many researchers utilize visual Simultaneous Localization and Mapping (SLAM) algorithm to address the reconstruction issue, like feature-based ORB-SLAM \cite{orbslam}, or direct approach like DSO \cite{dso}, or LSD-SLAM \cite{lsdslam} to predict depth. In these works, firstly the 6-DoF pose of camera is estimated from the monocular video stream, then a sparse or semi-dense map is reconstructed to obtain the corresponding depth information.

Compared to the traditional methods, recent approaches focus on casting depth estimation to an end-to-end problem using deep learning. Deep learning based depth estimation can also be divided into two categories: Supervised and unsupervised learning.

\subsection{Supervised Learning Methods}
In recent years, there have been a lot of work directed to the supervised learning methods, such as deeper depth prediction \cite{deeperdepth}, FCN \cite{fcn}, U-Net \cite{unet} and others. These approaches demonstrate that monocular depth prediction problem can be addressed in an end-to-end way using convolutional neural networks. But one thing that cannot be ignored is that the supervised learning is extremely dependent on a specific scene, on which the model is trained. Therefore making inference in another scenario may fail or give poor estimation results. Due to the aforementioned limitations, the researchers consider the unsupervised way of learning.

\subsection{Unsupervised Learning Methods}
Godard et al, proposed a self-supervised learning approach \cite{monodepth1}. The depth information estimated by the left image of the stereo camera is fine-tuned by the right image, which can be regarded as a geometric constraint of the left image. This way, the geometric information of the right frame is used to supervise the left frame. Although this approach utilizes the single-view image during the inference period, it requires a stereo image pair during the training phase, which is not available for using monocular image thoroughly. After all, we do not always have a stereo camera to perceive the surrounding environment.

Zhou et al. \cite{sfmleaner} proposed an algorithm to overcome the shortcomings of the aforementioned algorithm, such that even in the training phase, the single-view images are used instead of image pairs. The network architecture consists of a pose network and a disparity network. The pose network estimates the relative pose between the input frames. The disparity network estimates the disparity map of each input frames. With the predicted pose and the disparity map, we can reconstruct the target image. Then the photometric error is calculated between the reconstructed image and the original target image, which functions as the loss to train the neural network. In this way, the network combines 3D geometric information and 2D disparity information to estimate the depth value. A good depth can lead to a better pose estimate, and vice versa. In other words, good pose estimation and good depth estimation benefits from one another. The approaches proposed in vid2depth \cite{vid2depth}, struct2depth \cite{struct2depth}, GeoNet \cite{geonet} are based on the architecture of Zhou et al.

In addition to the reference model of Zhou et al., Mahjourian et al. \cite{vid2depth} have proposed a 3D loss function, where the iterative closest points (ICP) \cite{icp} algorithm and 3D point cloud are used. This approach also adds principled masking instead of a general-purpose learned mask to handle the projecting problem. Casser et al. \cite{struct2depth} has modified the network model of Zhou et al. and added a motion model to solve the occlusion problem in the monocular video stream. 

GASDA \cite{zhaoshanshan} is proposed in 2019 to estimate the depth map using the Generative Adversarial Networks (GANs). In the proposed network, the input image style is transferred to the KITTI-style image and then prediction of the depth is implemented in a supervised way. The author claims that this novel network can solve the domain shift problem, however such transferring style may fail when switching from the outdoor to the indoor suddenly.

GANVO \cite{ganvo} predicts the depth map based on the work DCGAN by Radford et al. \cite{dcgan}. The framework is like a general GAN, where the generator part generates the depth map and the discriminator tries to discriminate based on the depth map and the estimated pose whether the reconstructed image is the original input image or the fake one. Generator is trained to generate a depth map with a higher accuracy, which means it reconstructs an image that is as similar as possible to the target image. At the same time, discriminator is trained to differentiate the target image from the reconstructed image. Although this approach is novel, the training of the discriminator is not trivial. The hard point for the training of the discriminator lies in the absence of the corresponding ground truth of the target image. However, if we already have the ground truth of the target image, then calculating the loss between the target image and reconstructed image is straight-forward, and does not require further training. Moreover, in comparison with the state-of-the-art model such as monodepth2 \cite{monodepth2},  GANVO performance is worse.


\section{Method}
\subsection{Architecture}
\begin{figure}
    \centering
    \includegraphics[scale=0.3]{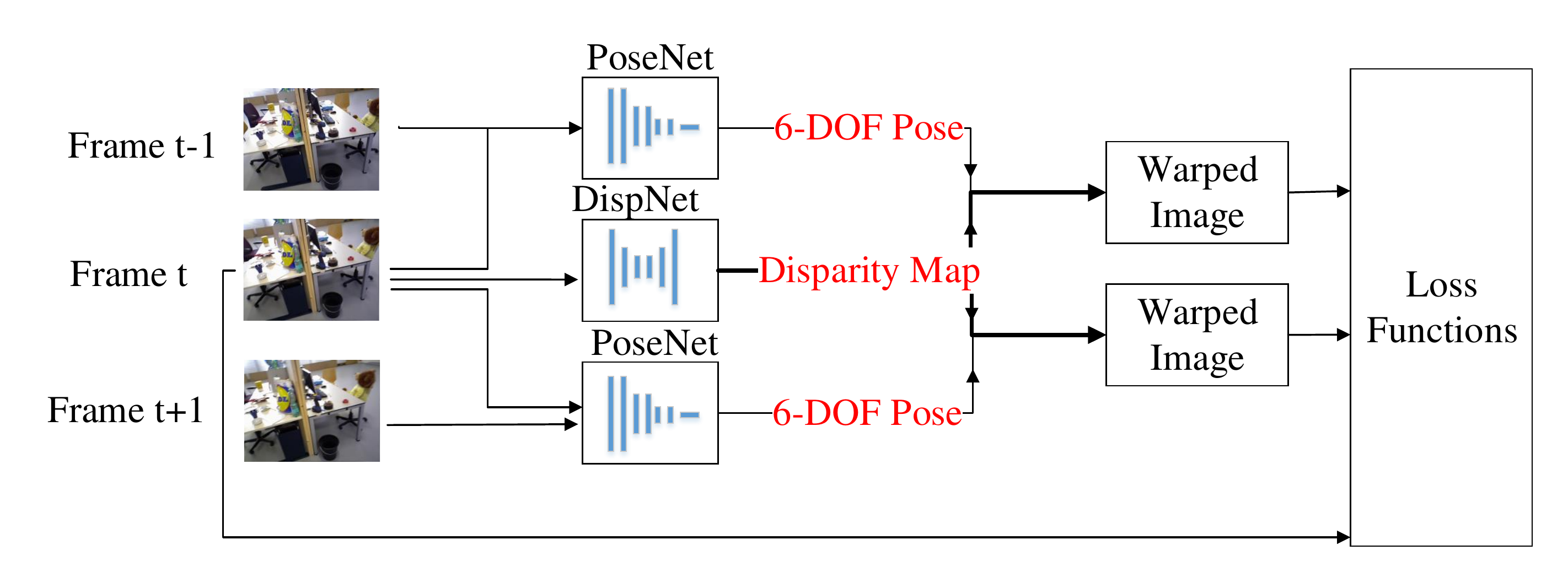}
    \caption{Applied Network architecture.}
    \label{fig:architecture_method1}
\end{figure}

Fig.~\ref{fig:architecture_method1} shows the neural network architecture of the  proposed method. We adopt the architecture from SfMLearner \cite{sfmleaner}, which is in turn based on DispNet\cite{dispnet}. The CNN consists of two towers: A depth tower receives the single-view image, and predicts the dense disparity map; a sequence of single-view images is fed into an ego-motion tower and the relative pose in an angle-axis form is generated.


\subsection{2D Losses}
Depth estimation is an essential problem in the field of 3D computer vision. To enable the neural network to learn the 3D information of the input images, it is necessary to consider the camera's 3D rigid motion, so the constraint of multi-view geometry is taken into account, when designing the loss function. First, the warping process is discussed. So we can warp the image $I_{t-1}$ based on $D_{t-1}$ and $T_{t}$,
\begin{equation}\label{warp}
   p^{warp}_{t-1}\sim KT_{t}D_{t-1}K^{-1}p_{t-1}
\end{equation}

\noindent where $p_{t-1}$ denotes the homogeneous coordinates of a pixel in the frame $t-1$, $K$ states the camera intrinsic matrix, $p_{t}$ represents the homogeneous coordinates of a pixel in the frame $t$, $T_{t}$ is the camera's movements from time $t-1$ to $t$ and $D_{t-1}$ is the estimated depth of frame $t-1$.

Then we use the differentiable bilinear sampling approach \cite{bilinearsampling}, to interpolate the values of the four neighboring pixels of $p_{t}$ to obtain the values of the warped image $I^{warp}_{p_{t}}$. The four-pixel includes the top left, top right, bottom left and bottom right. Finally the reconstruction loss is based on the photometric error:

\begin{equation} \label{recon_str}
    L_{reconstr}=\sum _{p_{t}\in \left\{ i,j\right\} }\left\| I_{p_{t}}-I^{warp}_{p_{t}}\right\|
\end{equation}

\begin{figure}
    \centering
    \subfigure[vid2depth \cite{vid2depth}]{
    \includegraphics[height=2.13cm]{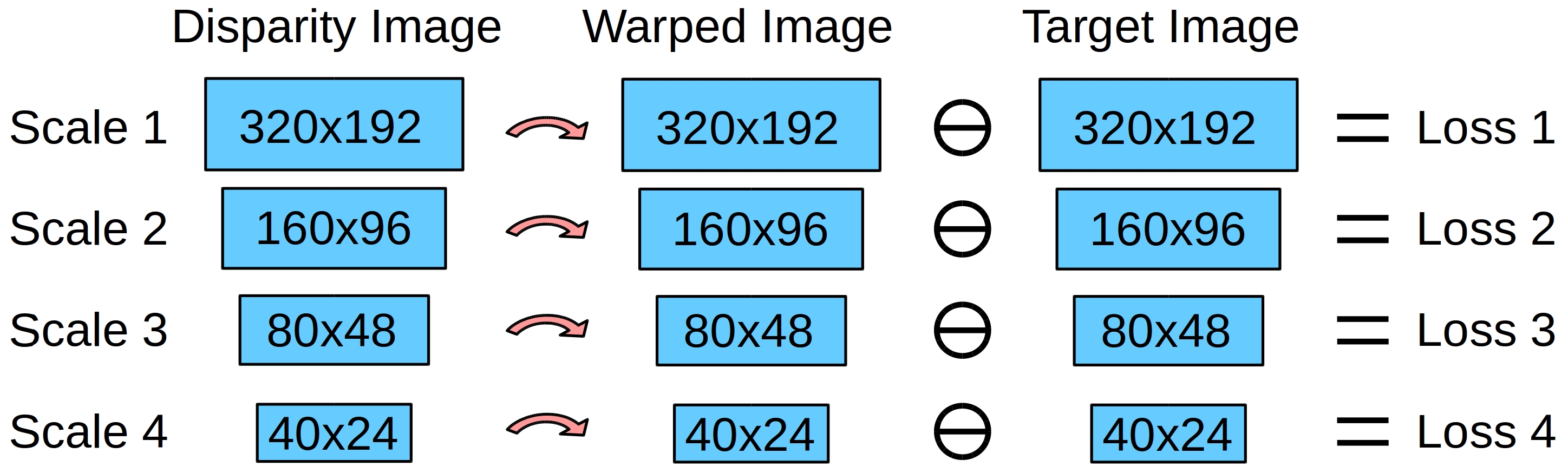}
    }
    \subfigure[our method]{
    \includegraphics[height=2.15cm]{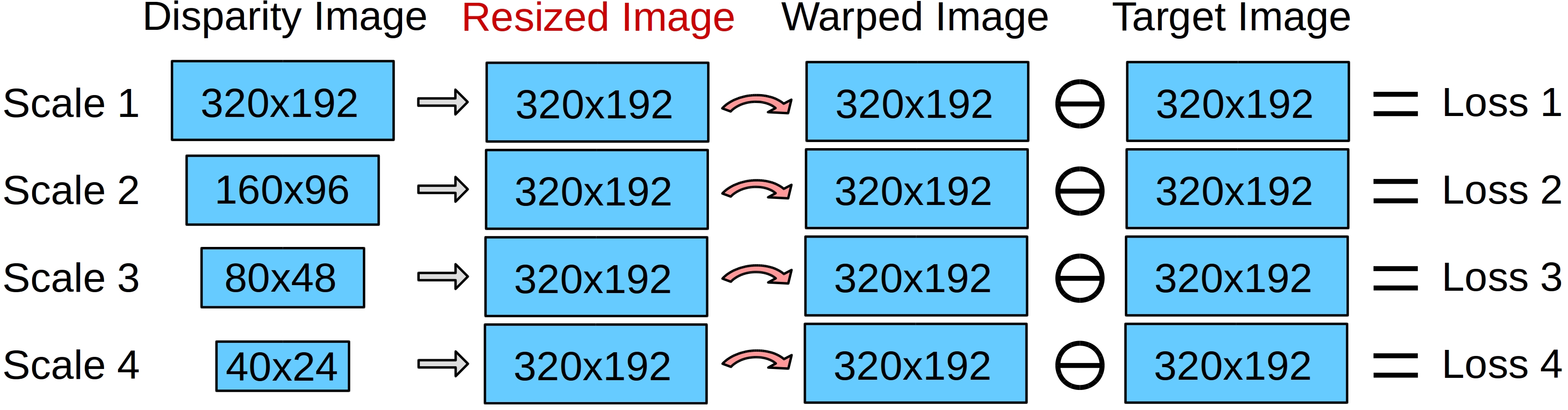}
    }
    \caption{Applied Loss Strategy.}
    \label{fig:modified_loss}
\end{figure}

In Eg. (\ref{warp}), $T_{t}$ is predicted by the PoseNet architecture, $D_{t-1}$ is generated by the DispNet architecture, which are shown in Fig. \ref{fig:architecture_method1}. In order to retain the depth information better, the DispNet generates multi-scale disparity map during the training phase. E.g., ${D}^{0}_{t-1}$ with original input size $320 \times 192$, ${D}^{1}_{t-1}$ with size $160\times96$, ${D}^{2}_{t-1}$ with size $80\times48$ and ${D}^{3}_{t-1}$ with size $40\times24$. 
 
As shown in Fig.~\ref{fig:modified_loss}~(a), the losses are designed basing on the image transformations, where the target image $I_{p_{t}}$ is resized along with the bilinear interpolation to create different sizes of warped images $I^{warp}_{p_{t}}$, which, in turn, are generated by warping the disparity maps of various sizes like $D^{0}_{t-1}$, $D^{1}_{t-1}$, $D^{2}_{t-1}$, $D^{3}_{t-1}$. And then  through $I_{p_{t}}$ and $I^{warp}_{p_{t}}$ we can obtain the reconstruction loss by Eq.~(\ref{recon_str}).

Inspired by monodepth2 and SfmLearner, our approach in computing the loss is shown in Fig.~\ref{fig:modified_loss}~(b). Unlike SfmLearner, we upsample the multi-scale disparity image to the input resolution, and then obtain the warped image $I^{warp}_{p_{t}}$. This means that instead of resizing the target image $I_{p_{t}}$ to different size, our approach transforms different sizes of the warped images $I^{warp}_{p_{t}}$ to the target image size. Then the reconstruction loss is built by the Eq.~(\ref{recon_str}). The advantage of this approach is that the converged reconstruction loss will be smaller than the method shown in Fig.~\ref{fig:modified_loss}~(a). The reason lies in the fact that the generated multi-scale disparity map is resized to equalize the input image's resolution, so that the error of the disparity map ($D^{i}_{t-1}, i\in{0,1,2,3}$) produced by DispNet in the first, second, third and fourth stages will be bounded, thus generating a more precise adjustment to each layer. Otherwise, with the approach shown in Fig.~\ref{fig:modified_loss}~(a),  the loss focuses on updating the weights of the last layer and ignores the details of the depth map, which results in the predicted depth map being not sharp enough. The similarity loss mentioned below is also computed according to Fig.~\ref{fig:modified_loss}~(b).

The structural similarity (SSIM) index is a measure of the similarity between two images. The SSIM ranges from zero to one. When two images are the same, the value should be one. The formula is defined as:

\begin{equation}\label{ssim}
    SSIM\left( x,y\right) =\dfrac {\left( 2\mu _{x}\mu _{y}+c_{1}\right) \left( 2\sigma_{xy}+c_{2}\right) }{\left(\mu_{x}^{2}+\mu_{y}^{2}+c_{1} \right) 
    \left( \sigma_{x}+\sigma_{y} + c_{2} \right) }
\end{equation}

where $\mu_{x}$, $\sigma_{x}$ are the local means and variances, respectively, and $x$, $y$ refer to two different images.\\

In this work, it utilizes Eq.~(\ref{ssim}) to measure the similarity between warped image $I^{warp}_{p_{t}}$ and the target image $I_{p_{t}}$. The final similarity loss is given as:
\begin{equation} \label{ssim_loss}
     L_{ssim}=\sum _{p_{t}\in \left\{ i,j\right\} }\left[1 - SSIM\left( I_{p_{t}}, I^{warp}_{p_{t}}\right)\right]
\end{equation}

where $I_{p_{t}}$ is the target image and $I^{warp}_{p_{t}}$ is the warped image from the target image. Similar to SfmLearner, we make use of edge-aware depth smoothness loss:

\begin{footnotesize}
\begin{equation} \label{smoothloss}
  L_{smooth}=\sum\left|\dfrac{\partial D_{t}}{\partial x} \right|exp \left( -\left | \dfrac{\partial I_{p_{t}}}{\partial x}\right | \right) +
  \left|\dfrac{\partial D_{t}}{\partial y} \right|exp\left( -\left | \dfrac{\partial I_{p_{t}}}{\partial y}\right | \right)
\end{equation}
\end{footnotesize}

where $\dfrac{\partial D_{t}}{\partial x}$ and $\dfrac{\partial D_{t}}{\partial y}$ are derivative of the depth map $D_{t}$ in the x-direction and y-direction, respectively, $\dfrac{\partial I_{p_{t}}}{\partial x}$ and $\dfrac{\partial I_{p_{t}}}{\partial y}$ are depth derivative of the image $I_{p_{t}}$ in the x-direction and y-direction, respectively.

\subsection{3D Loss}

Similar to vid2depth, the 3D loss is defined as:

\begin{equation} \label{3d_loss}
    L_{3D} = \left\| T^{'}_{t} - I \right \| + \left \| r_{t} \right \|
\end{equation}

\noindent where $\left\| . \right\|$ denotes the $L_{1}$ norm, $I$ is the identity matrix, $r_{t}$ is the regression point to error in the depth image $D_{t}$ and $T^{'}_{t}$ is the Iterative Closest Point (ICP) residual. In the end, 2D losses (SSIM loss and Reconstruction loss) are applied at same scale while 3D loss and smoothness loss functions are used at four different sizes of disparity image. The final loss is:
\begin{small}
\begin{equation} \label{final_loss}
    L_{final} = \sum \left( \alpha L_{reconstr} + \beta L_{ssim} \right) + \sum_{s} \left( \gamma L_{smooth}^{s} + \omega L_{3D}^{s} \right)
\end{equation}
\end{small}

\subsection{Predicted Disparity Map}
\begin{figure}
    \centering
    \subfigure[rgb image]{
    \includegraphics[width=3cm,height=2cm]{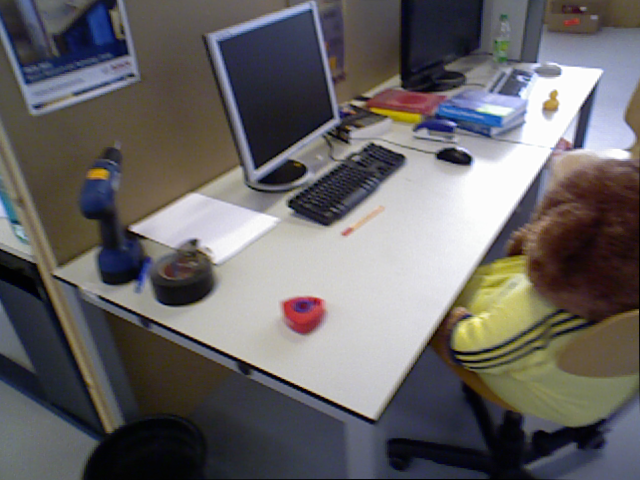}
    \includegraphics[width=3cm,height=2cm]{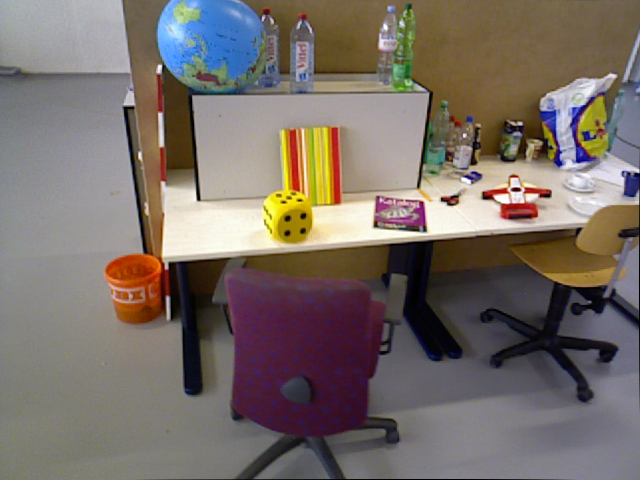}
    }
    \subfigure[ground truth]{
    \includegraphics[width=3cm,height=2cm]{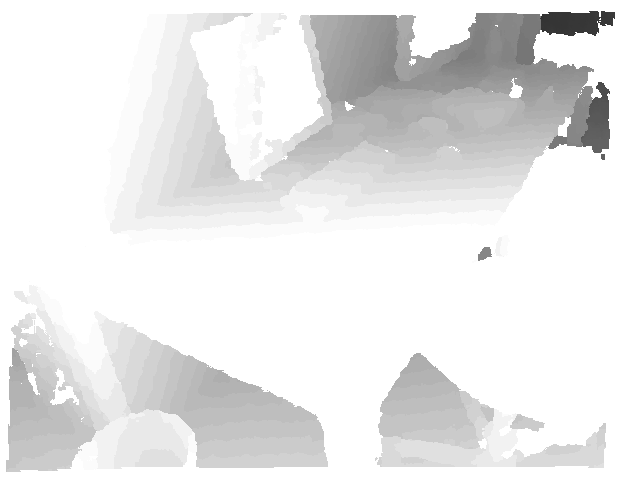}
    \includegraphics[width=3cm,height=2cm]{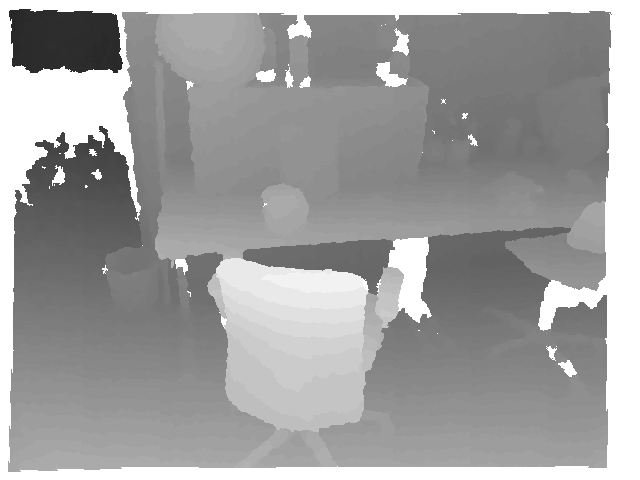}
    }
    \subfigure[vid2depth \cite{vid2depth}]{
    \includegraphics[width=3cm,height=2cm]{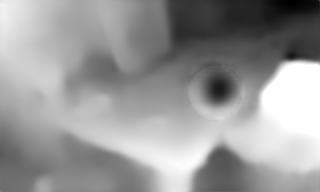}
    \includegraphics[width=3cm,height=2cm]{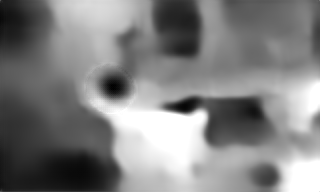}
    }
    \subfigure[our method]{
    \includegraphics[width=3cm,height=2cm]{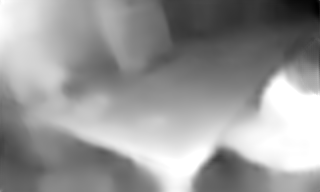}
    \includegraphics[width=3cm,height=2cm]{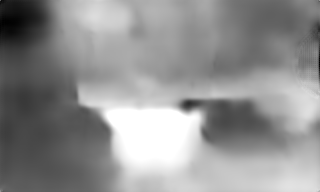}
    }
    \caption{Comparison of our method with vid2depth for depth prediction.}
    \label{fig:mothed_1_disparity_map}
\end{figure}

We use this method to train our model from scratch, and also train the model of vid2depth, which is the state-of-the-art unsupervised learning monocular depth prediction approach. As shown in Fig.~\ref{fig:mothed_1_disparity_map}, in comparison with vid2depth, our approach performs better since (i) it produces higher quality depth map with less noise compared to vid2depth and (ii) there are no black holes in resulting depth map as in vid2depth.

\section{General Improvements}
In order to make the estimated depth map more accurate, some techniques are discussed in this section. We are committed to exploring the possibility of improving the quality of the depth map, and thus two improvements are proposed: Ensemble learning with flipping strategy (ELWF) and median filtering.

\subsection{Ensemble Learning with Flipping Strategy}

\textbf{Post-processing} was utilized by Godard \cite{monodepth1} to reduce the occlusion effect caused by stereo image pairs. The left edge of the generated right-image disparity map and the left edge of the left-image disparity map have disparity ramps, and post-processing was proposed to overcome them. So firstly, they infer the input left image  $I_{left}$ to generate disparity map $I^{l}_{disp}$, and then left image is flipped $I^{flip}_{left}$ and inferred to produce the disparity map $I^{l_{f}}_{disp}$. After that the image $I^{l_{f}}_{disp}$ are flipped to generate $I^{l_{ff}}_{disp}$. The final image is $I^{l}_{disp_{final}}$. The 5\% left and right sides of $I^{l}_{disp_{final}}$ comes from 5\% left of $I^{l}_{disp}$ and 5\% right of $I^{l_{ff}}_{disp}$, respectively, while the middle part is the average of $I^{l}_{disp}$ and $I^{l_{ff}}_{disp}$.


\vspace{0.3cm}
\textbf{Ensemble Learning} \cite{ensamblelearning} \cite{ensemblelearning2} is a widely applied for nonlinear methods such as deep learning. The disadvantage of deep learning is that the inference results largely depend on the training data. In general, such neural networks with high variance can cause performance degradation. Ensemble learning reduces the model variance by training multiple networks rather than the individual ones and combine their results. This increases the prediction accuracy for applications with high variance. 



\begin{figure}
    \centering
    \includegraphics[width=1.0\linewidth]{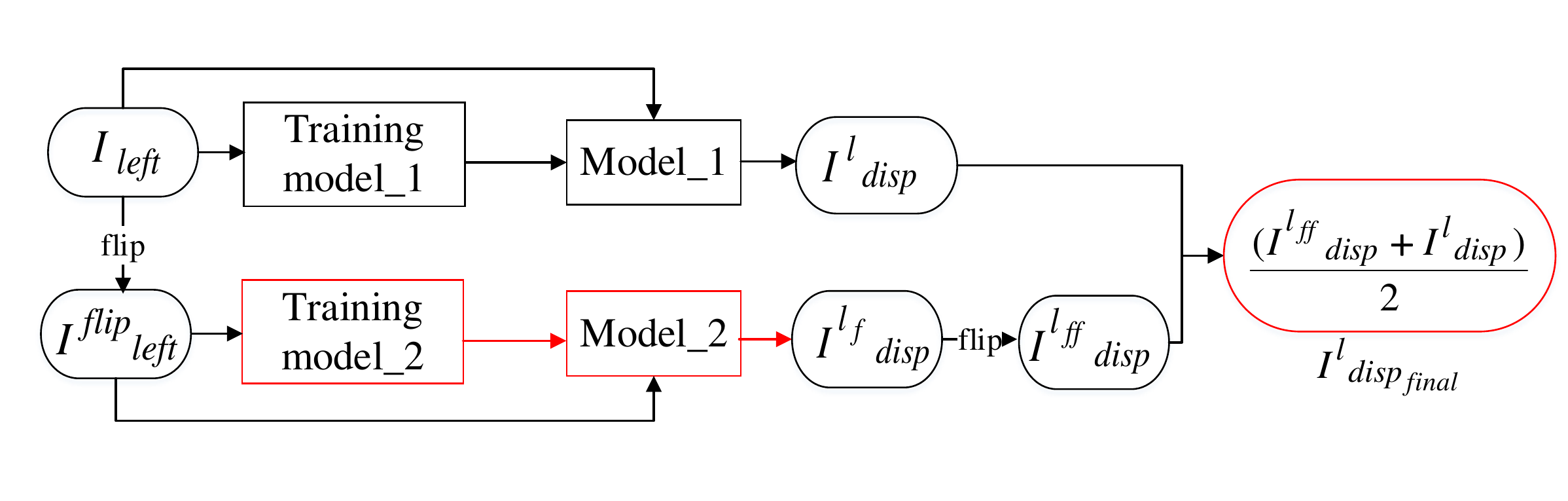}
    \caption{Ensemble Learning with Flipping Strategy (ELWF).}
    \label{fig:ensemble_learning_with_flipping_strategy}
\end{figure}

Inspired by the Post-Processing and ensemble learning, we propose a novel approach called the ensemble learning with flipping strategy (ELWF). Our approach is shown in Fig.~\ref{fig:ensemble_learning_with_flipping_strategy}. Unlike the post-processing above, the model is trained twice from scratch to obtain two models in the training phase: One training through unflipped images and one training with flipped images. Next, the disparity maps from the first model and the second model are inferred and the generated disparity maps are averaged to obtain the final disparity.

\begin{figure}
    \centering
    \includegraphics[width=3cm,height=2cm]{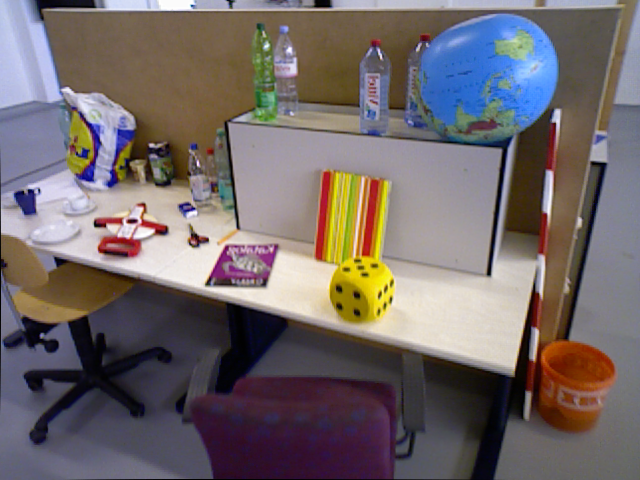}
    \includegraphics[width=3cm,height=2cm]{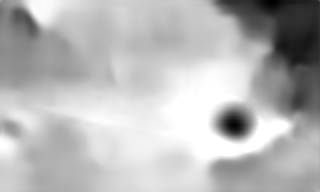}
    \caption{Poor quality disparity map produced by model that is trained with unflipped images but tested by flipped images.}
    \label{fig:poor_quality_disparity_map}
\end{figure}

The reason of flipping the input images and retraining the model from scratch is motivated by a part of the TUM dataset (freiburg3) \cite{tumrgbd}, where the camera rotates around the table from right to left in the video stream. Since PoseNet can only learn a left-turning pose, the network cannot adapt to the situation of rotating to the right, which means when the PoseNet receives an input video stream rotated to the right, the network still estimates a pose rotation to the left, resulting in an inferior quality of the estimated depth map as in Fig.~\ref{fig:poor_quality_disparity_map}.

The advantage of our approach is that after flipping the image and retraining the network, the final estimated disparity map can be balanced to avoid the occurrence of large errors in the estimated depth value, thus leading to a higher accuracy.

\begin{figure}
    \centering
    \subfigure[input image]{
    \includegraphics[width=3cm,height=2cm]{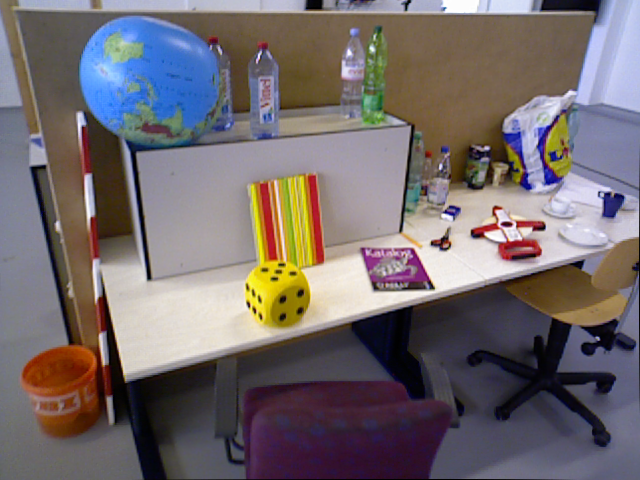}
    }
    \subfigure[trained model with un-flipped images]{
    \includegraphics[width=3cm,height=2cm]{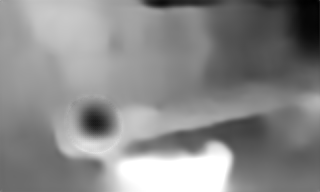}
    }
    \subfigure[trained model with flipped images]{
    \includegraphics[width=3cm,height=2cm]{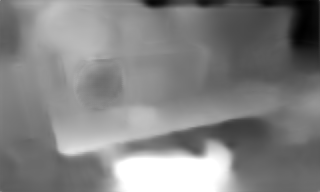}
    }
    \subfigure[average above]{
    \includegraphics[width=3cm,height=2cm]{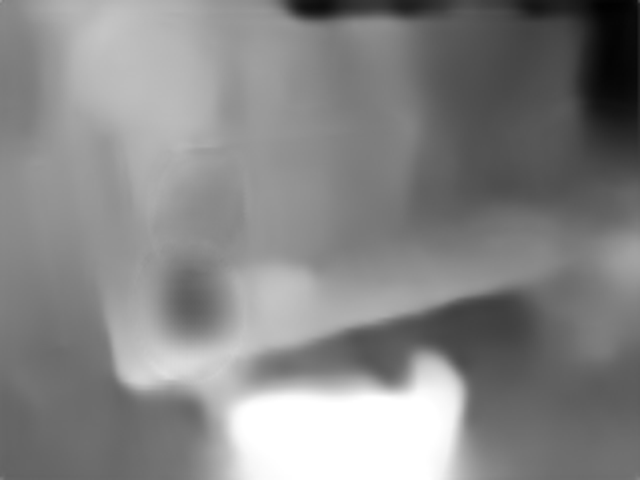}
    }
    \caption{Depth prediction through Ensemble Learning with Flipping (ELWF).}
    \label{fig:depth_prediction_ensemble_learning_with_flipping_strategy}
\end{figure}

\subsection{Filtering}
As stated by Liu et al.\cite{3D-sensing}, DL-based depth estimation only estimates the depth distribution rather than the actual depth values. Therefore, we believe that the produced depth maps can be adjusted by filtering techniques, which make the distribution of depth value more reasonable.

\begin{figure}
    \centering
        \begin{tikzpicture} [scale=0.8]
          \begin{axis}[
           legend pos=south east,
            ymin=0.5,ymax= 2.5,
            enlargelimits=false,
            enlargelimits=0.15,
            ylabel={RMSE [m]},
            symbolic x coords={a,b,c,d},
            xtick=data,
            nodes near coords, 
            nodes near coords align={vertical},
            x tick label style={rotate=0,anchor=north},
            legend style={at={(0.62,0.95)},anchor=north},
            ]
            \addplot coordinates {(a,1.067) 
                                   (b,0.947) 
                                    (c,1.028) 
                                    (d,0.981)
                                    };
           \addlegendentry{TUM datasets(freiburg3)}
            \addplot coordinates {(a,1.4378) 
                       (b,1.456) 
                        (c,1.419) 
                        (d,1.404)
                        };
           \addlegendentry{X-tion datasets}
          \end{axis}
        \end{tikzpicture}
    \caption{Evaluation of different filtering algorithms: (a) vid2depth \cite{vid2depth}; (b) vid2depth + max filter 15; (c) vid2depth + median filter 35; (d) vid2depth + median filter 55.}
    \label{fig:diff_filtering_resluts}
\end{figure}
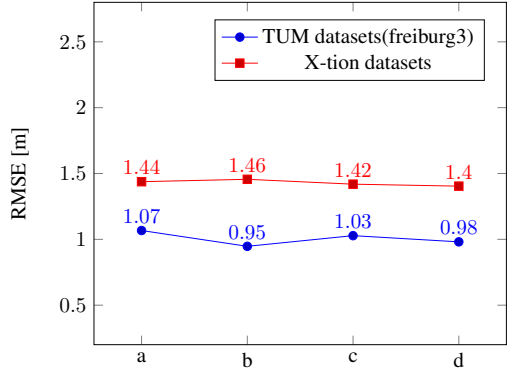

\begin{figure}
    \centering
    \subfigure[input image]{
    \includegraphics[width=3.5cm,height=2.5cm]{fig60000000160.png}
    }
    \subfigure[vid2depth \cite{vid2depth}]{
    \begin{tikzpicture}
    \node[anchor=south west,inner sep=0] (image) at (0,0) {\includegraphics[width=3.5cm,height=2.5cm]{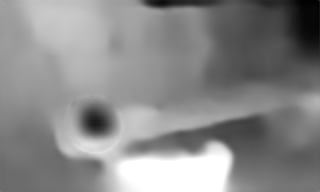}};
    \begin{scope}[x={(image.south east)},y={(image.north west)}]
    \draw[red,thick,dashed] (0.2,0.24) -- (0.4,0.24) -- (0.4,0.5) -- (0.2,0.5) -- (0.2,0.24);
    \end{scope}
    \end{tikzpicture}
    }
    \subfigure[vid2depth+ELWF]{
    \begin{tikzpicture}
    \node[anchor=south west,inner sep=0] (image) at (0,0) {\includegraphics[width=3.5cm,height=2.5cm]{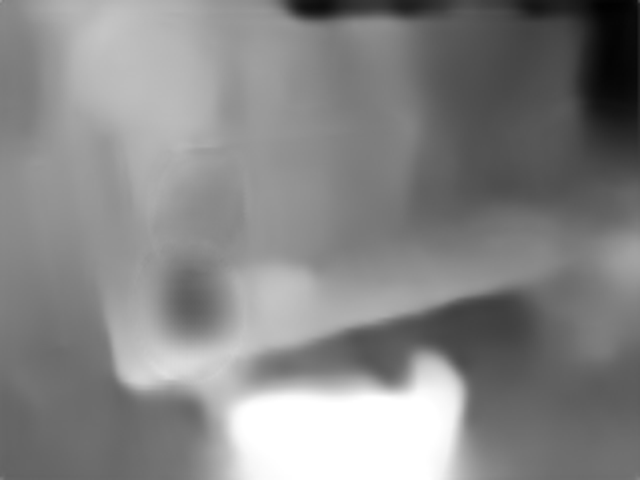}};
    \begin{scope}[x={(image.south east)},y={(image.north west)}]
    \draw[red,thick,dashed] (0.2,0.24) -- (0.4,0.24) -- (0.4,0.5) -- (0.2,0.5) -- (0.2,0.24);
    \end{scope}
    \end{tikzpicture}
    }
    \subfigure[vid2depth+ELWF+Filtering]{
        \begin{tikzpicture}
    \node[anchor=south west,inner sep=0] (image) at (0,0) {\includegraphics[width=3.5cm,height=2.5cm]{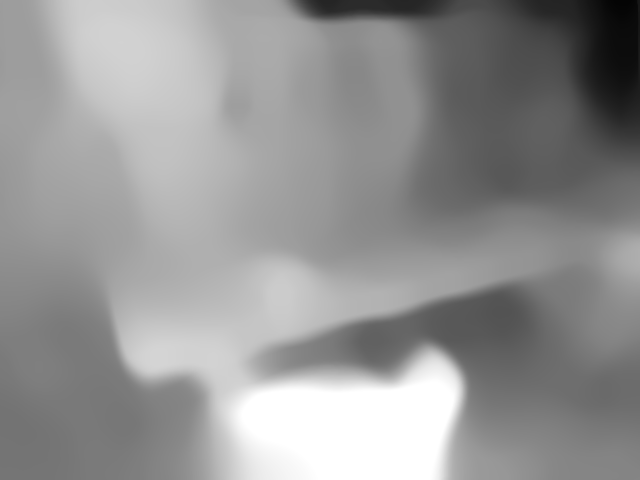}};
    \begin{scope}[x={(image.south east)},y={(image.north west)}]
    \draw[red,thick,dashed] (0.2,0.24) -- (0.4,0.24) -- (0.4,0.5) -- (0.2,0.5) -- (0.2,0.24);
    \end{scope}
    \end{tikzpicture}
    }
    \caption{The effect of ELWF and filtering over vid2depth.}
    \label{fig:before_and_after_filtering}
\end{figure}

In the implementation of the algorithm, we find that the estimated depth map has some 'black holes', as shown in Fig.~\ref{fig:before_and_after_filtering}~(b). The black pixel refers to the infinite distance. Since the estimated depth map is a distribution, we adopt the filtering algorithm to process the predicted depth image and improve the quality of our depth image. We compare the max filtering algorithm and the median filtering algorithm by applying different filter sizes to the estimated depth map, as shown in Fig.~\ref{fig:diff_filtering_resluts}. The red and blue lines in Fig.~\ref{fig:diff_filtering_resluts} denote results for X-tion and TUM RGB-D datasets, respectively. It has been observed that the median filter with the size of 35 performs well in both datasets. Although median filtering with filter size of 55 performs well, its filter size is very large, which will cause significant amount of depth information to be filtered out. Fig.~\ref{fig:before_and_after_filtering} shows that, in comparison to vid2depth, with the median filter and ELWF, the 'black holes' almost disappear and there is almost no noise in the depth map.

\section{ Implementation \& Experimental  Results}

\subsection{Implementation}
\subsubsection{TUM RGB-D Dataset}

We have used TUM RGB-D \cite{tumrgbd} dataset as the primary training and evaluation dataset. This dataset is the benchmark used for ego-motion accuracy evaluation, which includes the monocular video stream of a mobile robot and a single-view video stream of a handheld camera. This paper focuses on the monocular video of a hand-held SLAM under the folder fr3/long-office-household, which is the study of 10\% of handheld camera datasets. The reason we have chosen this TUM RGB-D dataset is as following:
\begin{enumerate}
    \item The image sequences contain rich texture information, which is more conducive to learning depth information by the deep neural network.
    \item The motion of the camera in this dataset is noticeable. This apparent motion facilitates the use of algorithms for motion estimation. For example, ORB-slam \cite{orbslam} and LSD-SLAM \cite{lsdslam} use this dataset to evaluate the performance of its Visual Odometry in the indoor monocular video streams. Since our depth estimation neural network needs to estimate the pose of the camera, this dataset is suitable for our work.
\end{enumerate}



The depth estimation in a monocular video stream highly relies on the motion between two adjacent frames of the camera. The baseline of the stereo camera is similar to the ego-motion between two adjacent frames of a monocular camera. When ego-motion is negligible, the baseline equivalent to the stereo camera is also small, which causes the scale uncertainty by the depth estimation. Assuming that the baseline is zero, the left and right camera of the stereo camera coincide, and the stereo camera becomes a monocular camera, in this case, the stereo camera cannot estimate the depth.

In general, with the same algorithm, a significant camera motion is preferred in estimating its pose. Therefore, in order to improve the accuracy of depth estimation, one frame is selected from every five frames. So 500 images are selected as the training data, and the remaining 1600 image is testing data. According to Zhou's method \cite{sfmleaner}, the input image sequence is pre-processed. To meet physical memory limitations of the machines that perform network training, we resize the input image resolution to $320 \times 192$ and change the format from 'png' to 'jpg'.

\subsubsection{Recorded Dataset}
X-tion is an RGB-D camera, with depth ranging from 0.8m to 3.5m, power consumption is less than 2.5 watts. Moreover, it has an Intel x86 architecture supportting OpenNI and its frame rate is 30 fps. In order to evaluate the proposed algorithm and determine whether the algorithm can estimate the depth map in the real scene, we have recorded an indoor scene video using X-tion camera and trained the network based on the recorded RGB images,. Afterwards we have used the depth images recorded by X-tion camera as the GT to evaluate our method. In total there are 1000 frames in X-tion dataset and 25\% of it used for training while the rest is used for testing. Different from TUM-RGB-D dataset, the videos are recorded by moving the camera straight forward. So that, we can evaluate our method with different camera motion using X-tion dataset.

\subsubsection{Training Details}
Our deep neural network models are implemented with the Tensorflow framework  \cite{tensorflow} and trained with one NVIDIA Quadro K5200 GPU. Models are trained for 627 epochs from scratch, and checkpoints are saved every 1000 epochs. We have used a computer equipped with NVIDIA GForce GTX1060 and Intel Core i7 CPU for testing. Adam optimizer is used with hyper-parameters $\alpha = 0.85$, $\beta = 0.15$, $\gamma = 0.15$, $\omega = 0.1$ and with learning rate of $2 \times 10^{-4}$.

\subsection{Experimental Results}
\subsubsection{Competing Approaches}
Monodepth1, monodepth2, vid2depth, struct2depth and sfmlearner are the state-of-the-art approaches for the unsupervised monocular depth prediction in the outdoor scenario, which performed well on the KITTI and CityScape datasets. However, in the indoor scenario such as TUM RGGB-D dataset, when these models are trained from scratch and their results are evaluated, we have observed that monodepth1 generates the white depth images, which means that the depth value is zero. Moreover, monodepth2, struct2depth and sfmlearner also produce unreasonable depth maps. Only vid2depth is able to get reasonable depth maps after training from scratch as shown in Fig.\ref{fig:before_and_after_filtering}~(b).

\begin{figure}
    \centering
    \subfigure[input images]{
    \includegraphics [width=2.8cm,height=2.1cm]{fig60000000160.png}
    \includegraphics [width=2.8cm,height=2.1cm]{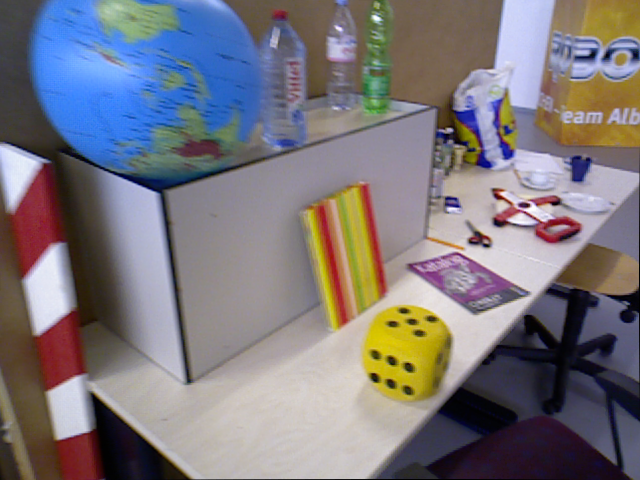}
    }
    \subfigure[sfmlearner \cite{sfmleaner} (train from scratch)]{
    \includegraphics [width=2.8cm,height=2.1cm]{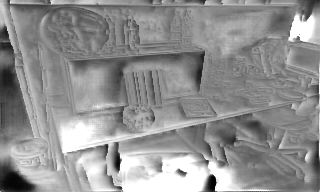}
    \includegraphics [width=2.8cm,height=2.1cm]{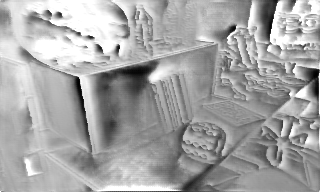}
    }
    \subfigure[monodepth2 \cite{monodepth2} (train from pretrained model)]{
    \includegraphics [width=2.8cm,height=2.1cm]{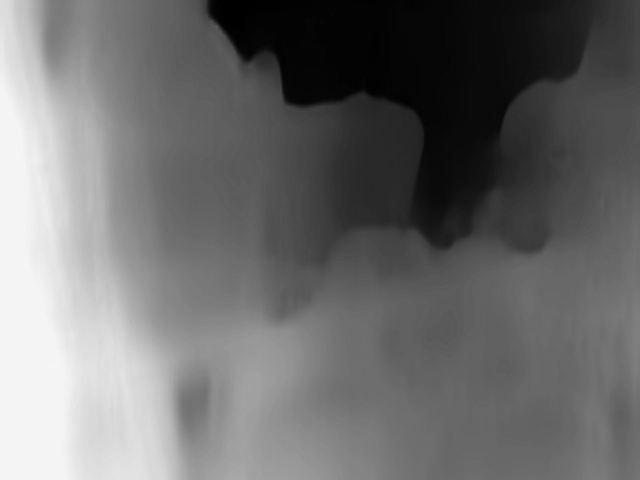}
    \includegraphics [width=2.8cm,height=2.1cm]{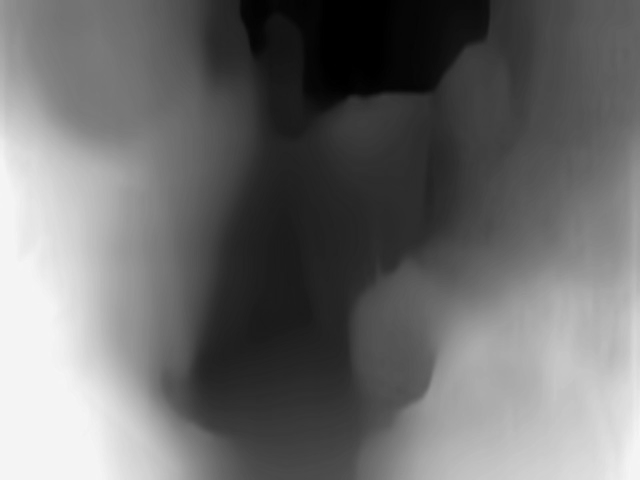}
    }
    \subfigure[struct2depth \cite{struct2depth} (train from pretrained model)]{
    \includegraphics [width=2.8cm,height=2.1cm]{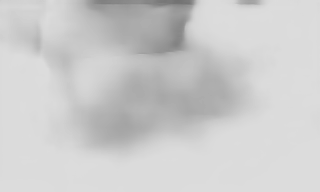}
    \includegraphics [width=2.8cm,height=2.1cm]{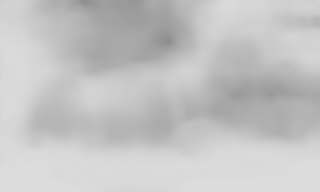}
    }
  
    \caption{Performance of sfmlearner, monodepth and struc2depth for indoor scenario.}
    \label{fig:sfmlearnerandstruc2depthresults}
\end{figure}

We have also fine tuned the models from pretrained models. Fig.\ref{fig:sfmlearnerandstruc2depthresults}~(c) shows that monodepth2 can not estimate the depth map correctly, e.g. the outlines of the table and the globe in the image are not estimated. For struct2depth, after training 350 epochs starting from pretrained model, the depth map becomes zero, as in Fig.\ref{fig:sfmlearnerandstruc2depthresults}~(d). In summary, Fig.\ref{fig:sfmlearnerandstruc2depthresults} and shows that sfmlearner, monodepth2 and struct2depth can not estimate the depth map in the indoor video streams, but vid2depth can get the depth map. So the main competing approach for our work is the vid2depth method.

\begin{table*}
\centering
\caption{Depth estimation results on TUM RGB-D dataset.}
\begin{adjustbox}{center}
\begin{tabular}[t]{cccccccc}
\hline
Method                          & Datasets           & RMSE       & Abs Rel      & Sq Rel   & $\delta < 1.25$ & $\delta < 1.25^{2}$ & $\delta < 1.25^{3}$  \\ 
\hline
vid2depth \cite{vid2depth}       &  TUM-freiburg3      &    1.067    &  0.308       & 0.915  & 0.710      & 0.710    & 0.710  \\
\hline
vid2depth+ELWF                  &  TUM-freiburg3      &    0.910    &  0.275       & 0.649    &  0.711    &  0.711   &  0.711  \\
vid2depth+ELWF+Median Filter    &  TUM-freiburg3      &    0.877    &  0.268       & 0.604    & \textbf{0.715}      & \textbf{0.715}    & \textbf{0.715}   \\
Our Method                     &  TUM-freiburg3      &    0.920    &  0.285       & 0.656    & 0.697      & 0.697    & 0.697  \\
Our Method  +ELWF              &  TUM-freiburg3      &    0.837    & 0.270        & 0.564    & 0.703       & 0.703    & 0.703 \\
Our Method +ELWF+Median Filter&  TUM-freiburg3      &    \textbf{0.812}   &  \textbf{0.264}       & \textbf{0.527} & 0.707   & 0.707  & 0.707\\
\hline
\end{tabular}
\end{adjustbox}
\label{tableresults1}
\end{table*}

\begin{table*}
\centering
\caption{Depth estimation results on X-tion dataset.}
\begin{adjustbox}{center}
\begin{tabular}[t]{cccccccc}
\hline
Method                          & Datasets           & RMSE       & Abs Rel      & Sq Rel   & $\delta < 1.25$ & $\delta < 1.25^{2}$ & $\delta < 1.25^{3}$  \\ 
\hline
vid2depth\cite{vid2depth}                &  X-tion data         &    1.475    &  0.217       & 0.450 & 0.688  & 0.688  & 0.688     \\
\hline
Our method                          &  X-tion data         &    \textbf{1.402}    &  0.206       & \textbf{0.400} & 0.712  & 0.712  & 0.712     \\
Our method + ELWF+ Median Filter    &  X-tion data         &    1.412    &  \textbf{0.200}       & 0.409 & \textbf{0.726}  & \textbf{0.726}  &\textbf{0.726}     \\
\hline
\end{tabular}
\end{adjustbox}
\label{taleX-tionresluts}
\end{table*}

\subsubsection{Depth Prediction Evaluation}

We have compared vid2depth, 'our method', 'our method + ELWF' and 'our method + ELWF + Median Filtering', whose visual examples are shown in Fig.~\ref{fig:depth_prediction_ensemble_learning_with_flipping_strategy}, respectively.  Our approach effectively avoids the noise or black hole, which are frequently observed in vid2depth. With the proposed ELWF median filtering, the produced depth map becomes sharper and of better quality. This is also reflected to the results acquired for TUM RGB-D dataset given in Table~\ref{tableresults1}. Compared to vid2depth, RMSE drops from 1.067 to 0.920 and Abs\_Rel decreases from 0.308 to 0.285 with our method.  With ELWF and median filtering, the above errors can be further reduced to 0.812 and 0.264, respectively. Consequently, our method significantly outperforms vid2depth on the task of unsupervised depth estimation with an improvement of 23.9\% and 14.3\% on RMSE and Abs\_Rel, respectively.

The vid2depth combined with ELWF and median filter also achieves better results than vid2depth alone. Abs\_Rel decreases from 0.308 to 0.268 and  RMSE drops from 1.067 to 0.877. Consequently, the quality of the predicted depth map has improved. This validates the applied ELWF and median filtering approaces, which improve the accuracy of indoor depth prediction even for different methods and have practical application value. 

Table~\ref{taleX-tionresluts} shows the results for the X-tion dataset. The models are trained on the X-tion-dataset from scratch. Compared to vid2depth, RMSE drops from 1.475 to 1.402 and Abs\_Rel drops from 0.217 to 0.206 with our method. So our method is a feasible model both in TUM-RGB-D and X-tion datasets. However, using ELWF does not guarantee the reduction of the error for all datasets. For example, in X-tion-dataset RMSE increases from 1.402 to 1.412 when ELFW + median filter setup is used. On the other hand, ELWF can reduce the RMSE error in the TUM RGB-D dataset. This phenomenon can be explained by the fact that in the TUM data, the camera always moves to the left around the table. With ELFW technique, our method learns both rotations (i.e. to the left and to the right) since flipped images are also used in the training. However, in the X-tion dataset, our camera moves straight forward in the room. When the image is flipped, the camera still moves forward without changing its motion, so the estimated motion of the model before and after the flipping is the same. Therefore, ELWF does not reduce the error on the X-tion dataset.

\begin{figure}[b]
    \centering
    \subfigure[RGB images]{
    \includegraphics[width=2cm,height=1.44cm]{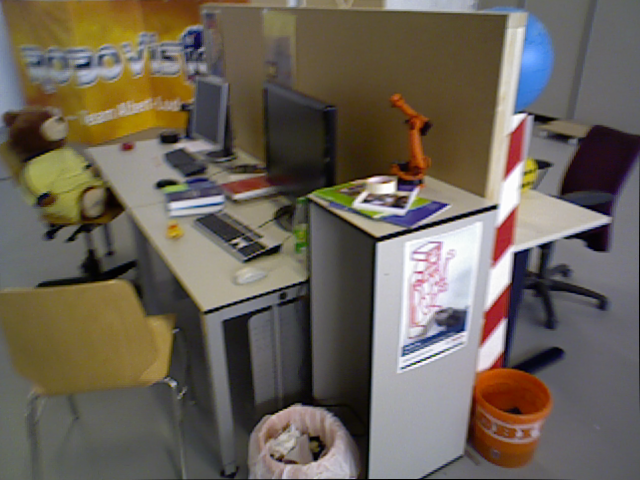}
    \includegraphics[width=2cm,height=1.44cm]{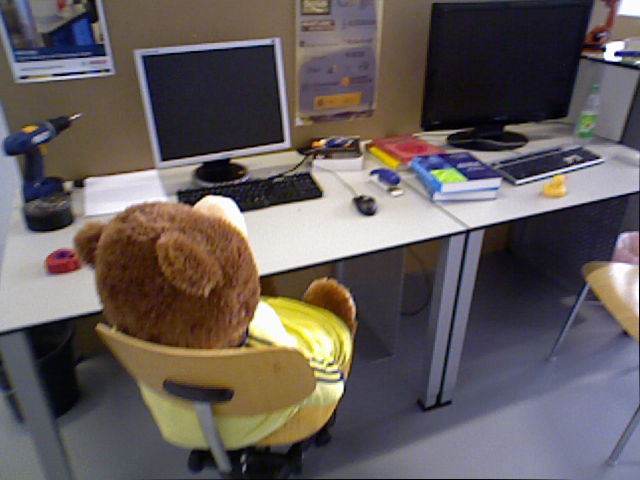}
    \includegraphics[width=2cm,height=1.44cm]{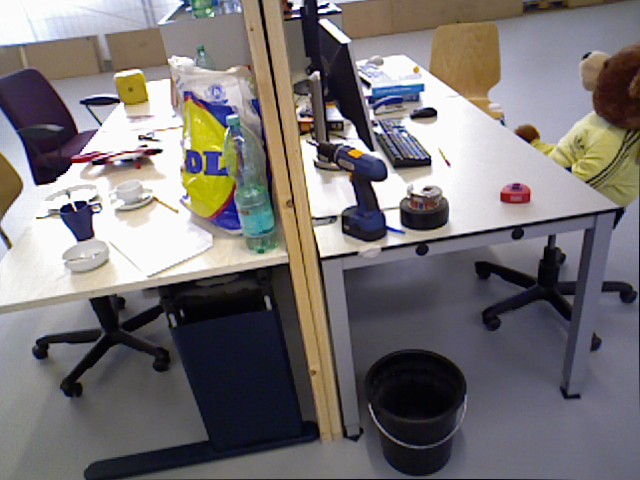}
    \includegraphics[width=2cm,height=1.44cm]{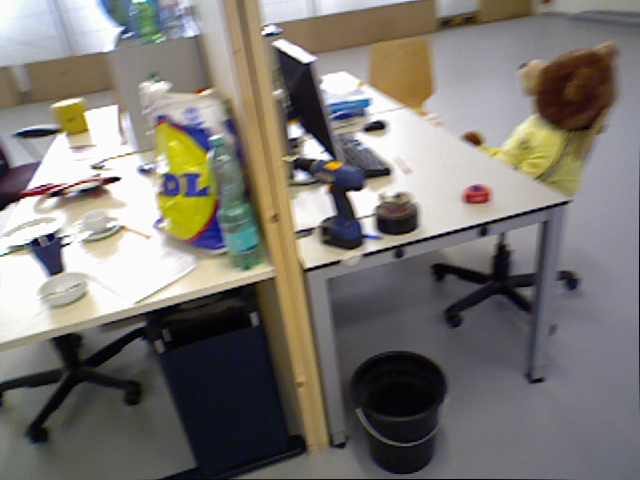}
    }
    \subfigure[ground truth]{
    \includegraphics[width=2cm,height=1.44cm]{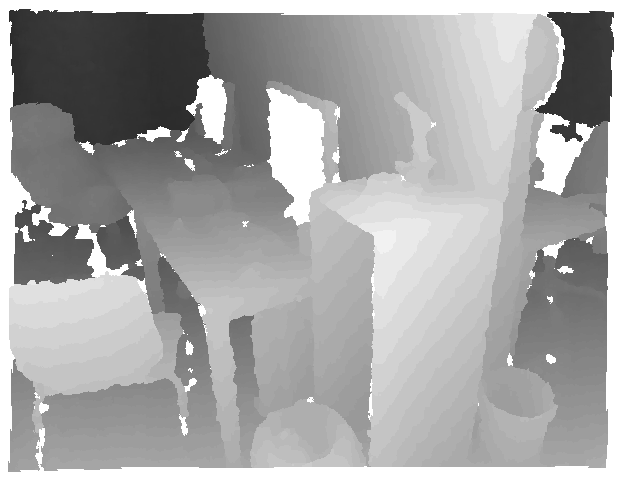}
    \includegraphics[width=2cm,height=1.44cm]{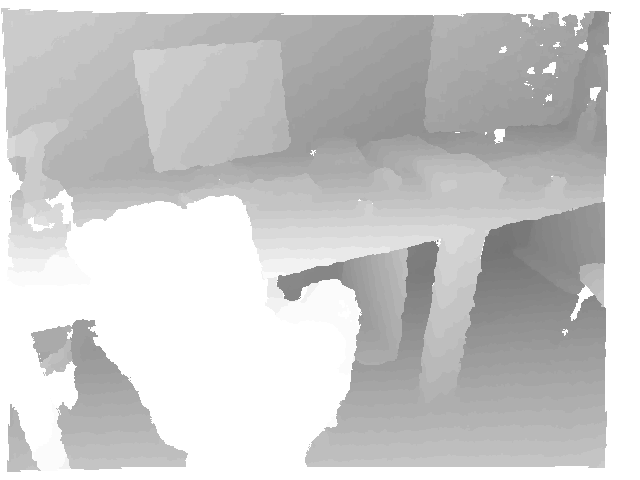}
    \includegraphics[width=2cm,height=1.44cm]{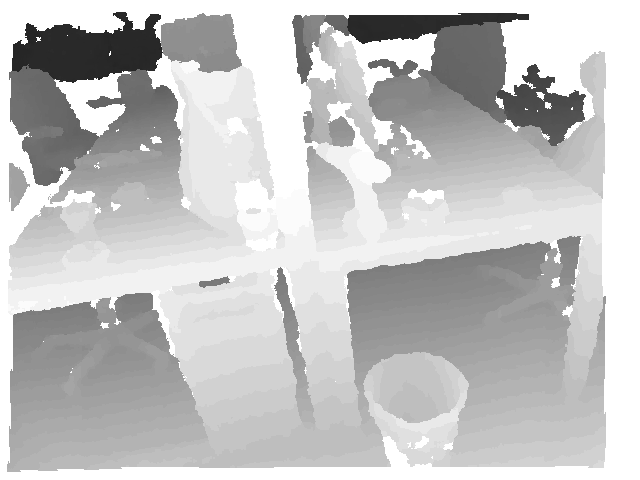}
    \includegraphics[width=2cm,height=1.44cm]{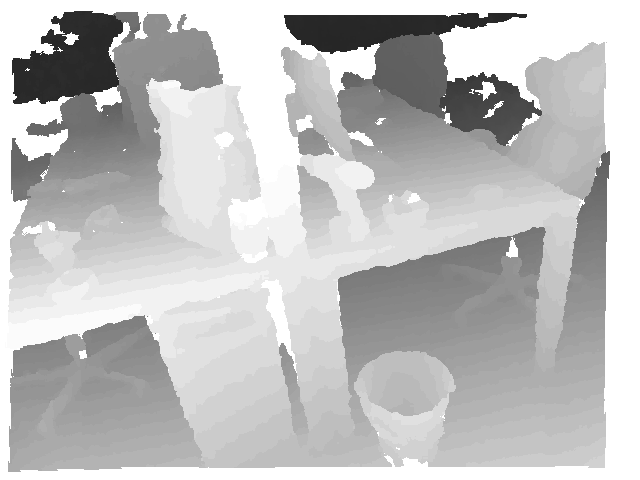}
    }
    \subfigure[vid2depth]{
    \includegraphics[width=2cm,height=1.44cm]{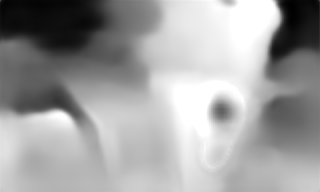}
    \includegraphics[width=2cm,height=1.44cm]{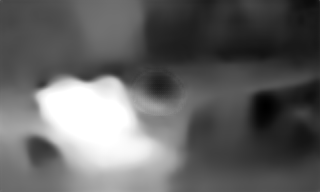}
    \includegraphics[width=2cm,height=1.44cm]{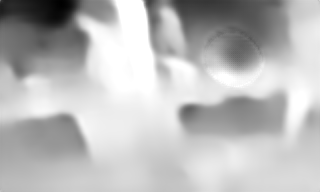}
    \includegraphics[width=2cm,height=1.44cm]{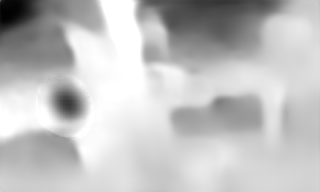}
    }
    \subfigure[our method ] {
    \includegraphics[width=2cm,height=1.44cm]{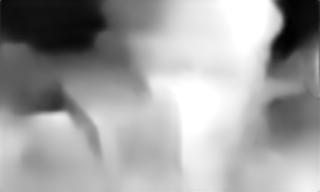}
    \includegraphics[width=2cm,height=1.44cm]{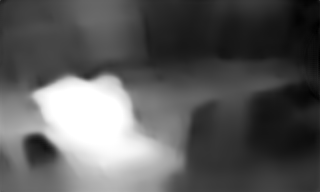}
    \includegraphics[width=2cm,height=1.44cm]{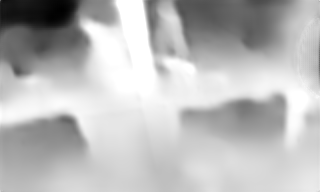}
    \includegraphics[width=2cm,height=1.44cm]{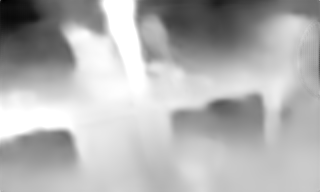}
    }
    \subfigure[our method + ELWF]{
    \includegraphics[width=2cm,height=1.44cm]{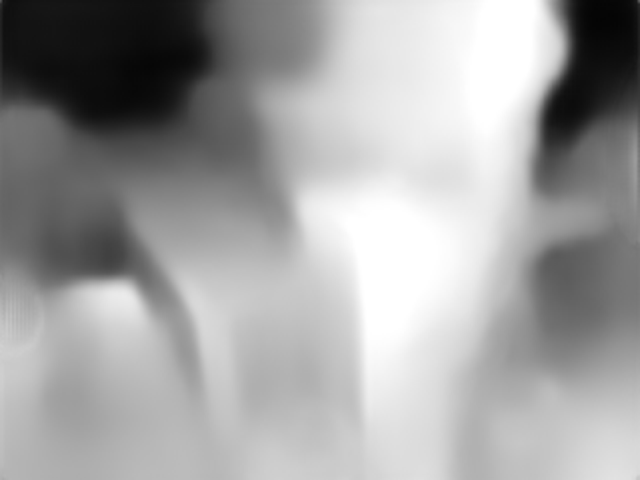}
    \includegraphics[width=2cm,height=1.44cm]{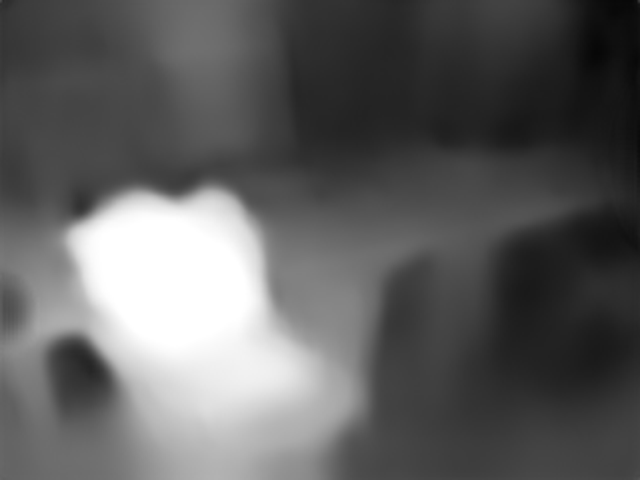}
    \includegraphics[width=2cm,height=1.44cm]{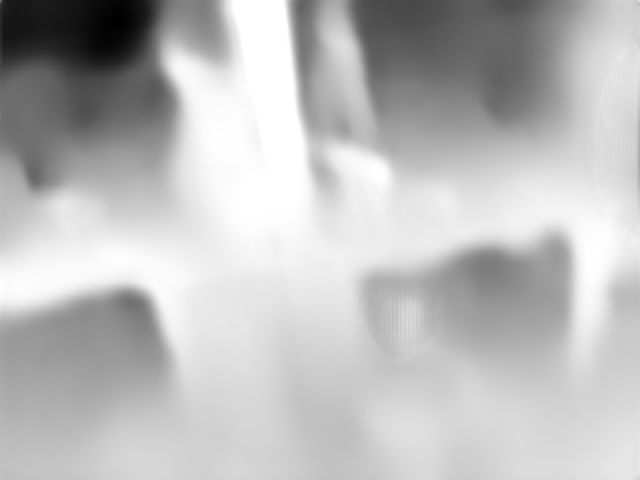}
    \includegraphics[width=2cm,height=1.44cm]{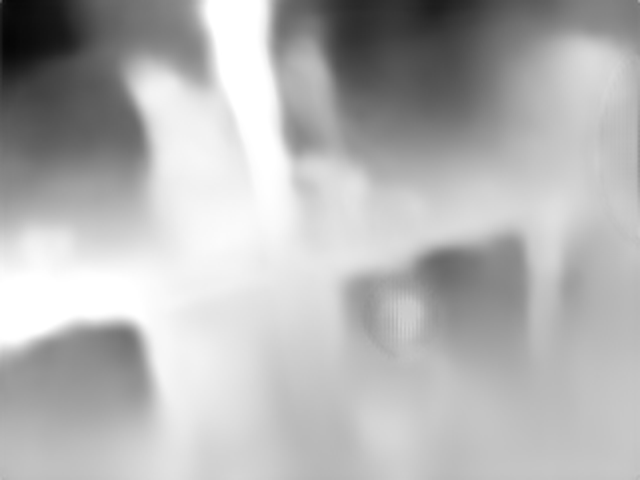}
    
    }
    \subfigure[our method + ELWF + Median Filter]{
    \includegraphics[width=2cm,height=1.44cm]{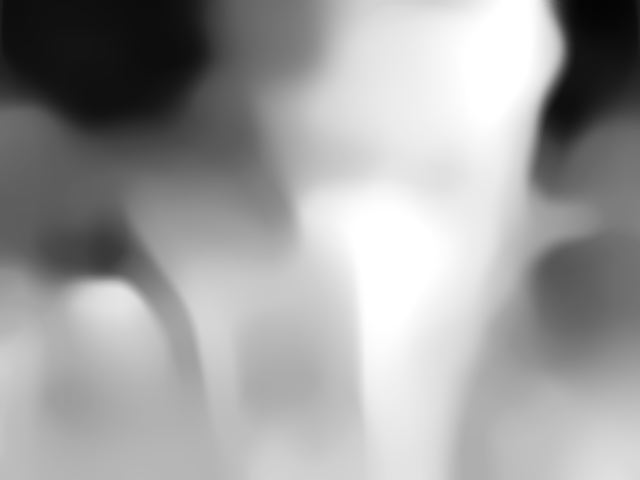}
    \includegraphics[width=2cm,height=1.44cm]{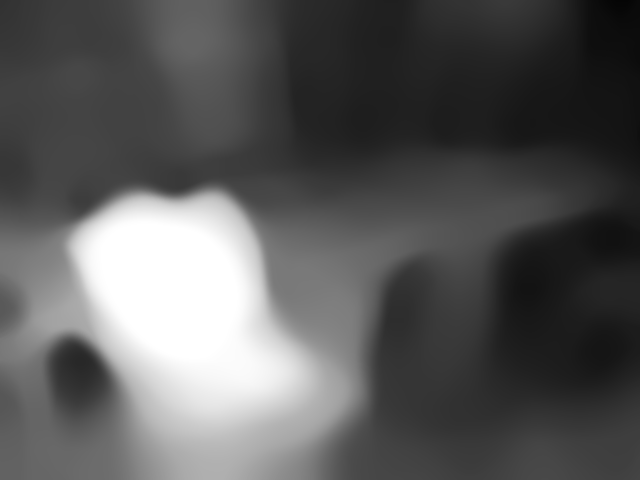}
    \includegraphics[width=2cm,height=1.44cm]{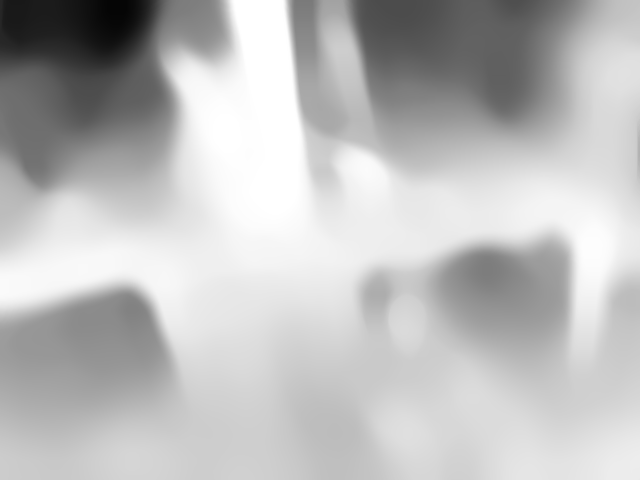}
    \includegraphics[width=2cm,height=1.44cm]{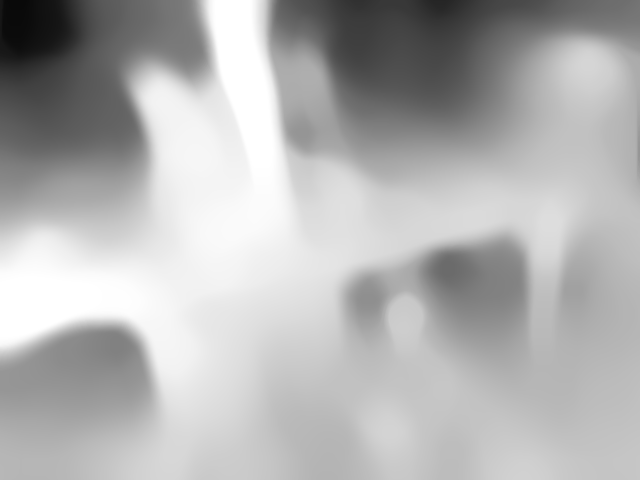}
    }
    \caption{Depth prediction performance of our method with ELWF and median filtering compared to vid2depth.}
    \label{fig:depth_prediction_ensemble_learning_with_flipping_strategy}
\end{figure}

\section{Conclusion}
In this paper, an unsupervised deep learning approach is proposed to estimate the depth maps for indoor monocular video streams. Compared to the outdoor scenes, the camera motion in the indoor scenes is more complicated. Therefore, we have analyzed the latest unsupervised depth estimation methods for indoor depth map estimation. Based on this, we propose our method to resize the generated disparity maps with varying scales to a uniform scale and to compute the 2D loss. This modification effectively helps the depth estimation algorithm to reduce the Abs\_Rel from 0.308 to 0.285 in the TUM dataset, and the RMSE decreases from 1.067 to 0.920. Moreover, with the proposed ELWF and median filtering, the errors are further reduced. Compared to the vid2depth method, the Abs\_Rel and RMSE are reduced by 14.3\% and 23.9\%, respectively. The experimental results verify that our method is practical and feasible.

As a future work, we plan to improve pose neural network in order to make the pose estimation more accurate and help to estimate a better depth map. Since the estimated depth is the distribution rather than the real depth value, we consider adding probability information to help the neural network estimate a better depth map.

{\small
\bibliographystyle{ieeetr}
\bibliography{egbib}
}

\end{document}